\documentclass[letterpaper, 10 pt, conference]{ieeeconf}
\IEEEoverridecommandlockouts
\usepackage{listings}
\usepackage{color}

\makeatletter
\let\NAT@parse\undefined
\makeatother
\usepackage[hidelinks]{hyperref}

\usepackage{amsmath}
\usepackage{amssymb}
\usepackage{amsthm}
\usepackage{subfig}
\usepackage{balance}
\usepackage{stfloats}
\usepackage{graphicx}
\usepackage{epstopdf}
\usepackage{bm}
\usepackage{booktabs}
\usepackage{array}
\usepackage{stfloats}
\usepackage{pdflscape}
\usepackage{pgfplots}
\usepackage{algorithm}
\usepackage{algorithmic}
\usepackage{setspace}
\usepackage{tikz}

\usepackage{cite}

\theoremstyle{plain}

\theoremstyle{definition}

\newtheorem{innercustomprb}{Problem}
\newenvironment{customprb}[1]
  {\renewcommand\theinnercustomprb{#1}\innercustomprb}
  {\endinnercustomprb}

\newtheorem{definition}{Definition}

\definecolor{mygray}{rgb}{0.5,0.5,0.5}
\definecolor{keyword}{rgb}{0.5,0.5,0.5}
\definecolor{greenCode}{rgb}{0, 0.6, 0}

\lstdefinelanguage{HTTP}{
  keywords={GET},
  ndkeywords={PUT},
  comment=[s]{PO}{T},
  morecomment=[s]{D}{LETE}
}

\lstdefinestyle{customc}{
  belowcaptionskip=1\baselineskip,
  language={HTTP},
  breaklines=true,
  frame=tb,
  captionpos=b,
  keywordstyle=\bfseries\color{greenCode},
  ndkeywordstyle=\bfseries\color{red},
  commentstyle=\bfseries\color{magenta},
  stringstyle=\bfseries\color{black},
  xleftmargin={0.75cm},
  showstringspaces=false,
  basicstyle=\footnotesize\ttfamily,
  numbers=left,
  numberstyle=\small\color{black},
}

\usepackage{flushend} 
\usepackage[colorinlistoftodos]{todonotes}

\usepackage{framed}

\makeatletter
\newcommand\fs@spaceruled{\def\@fs@cfont{\bfseries}\let\@fs@capt\floatc@ruled
  \def\@fs@pre{\vspace{0.5\baselineskip}\hrule height.8pt depth0pt \kern2pt}%
  \def\@fs@post{\kern2pt\hrule\relax}%
  \def\@fs@mid{\kern2pt\hrule\kern2pt}%
  \let\@fs@iftopcapt\iftrue}
\makeatother

\begin{document}
\title{\LARGE \bf CBGL: Fast Monte Carlo Passive Global Localisation \\ of 2D LIDAR Sensor}

\author{Alexandros Filotheou%
 \thanks{The author is with the Department of Electrical and Computer Engineering,
  Aristotle University of Thessaloniki, 54124 Thessaloniki, Greece  {\tt\small alefilot@auth.gr}}%
}

\maketitle
\thispagestyle{empty}
\pagestyle{empty}

\begin{abstract}
  Navigation of a mobile robot is conditioned on the knowledge of its pose. In
observer-based localisation configurations its initial pose may not be knowable
in advance, leading to the need of its estimation. Solutions to the problem of
global localisation are either robust against noise and environment
arbitrariness but require motion and time, which may (need to) be economised
on, or require minimal estimation time but assume environmental structure, may
be sensitive to noise, and demand preprocessing and tuning. This article
proposes a method that retains the strengths and avoids the weaknesses of the
two approaches. The method leverages properties of the Cumulative Absolute
Error per Ray (CAER) metric with respect to the errors of pose hypotheses of a
2D LIDAR sensor, and utilises scan--to--map-scan matching for fine(r) pose
estimations. A large number of tests, in real and simulated conditions,
involving disparate environments and sensor properties, illustrate that the
proposed method outperforms state-of-the-art methods of both classes of
solutions in terms of pose discovery rate and execution time. The source code
is available for download.

\end{abstract}

\begin{keywords}
global localisation, 2D LIDAR, monte carlo, scan--to--map-scan matching
\end{keywords}

\section{Introduction}
  The knowledge of a mobile robot's initial pose is a prerequisite in tasks
involving its navigation, especially in contexts where an observer is used for
pose tracking. However, the robot's pose is not predictable or pre-settable in
all conditions. This lack of predictability necessitates ad hoc estimation
without prior information. Various sensor and map modalities have been
investigated in the literature: from 2D and 3D LIDAR sensors
\cite{als_eth,Cop2018a}, RGBD cameras \cite{Guo2016}, and RFID equipment
\cite{Tzitzis2023b}, to keyframe-based submaps \cite{Lowry2016} and metric maps
\cite{Rosen2021}. In practice the latter combined with 2D LIDAR sensors have
become the de facto means of mobile robot localisation and navigation due to
the sensor's high measurement precision and frequency, almost no need for
preprocessing, and low cost compared to 3D LIDAR sensors.

This article addresses the problem of Passive Global Localisation of a 2D LIDAR
sensor in a 2D metric map, i.e. the estimation of its location and orientation
within the map, under complete locational and orientational uncertainty,
without prescribing robot motion commands
for further knowledge acquisition. The problem is
formalised in Problem \ref{prob:the_problem}:

\begin{customprb}{P}
  \label{prob:the_problem}
  Let the unknown pose of an immobile 2D range sensor whose angular range is
  $\lambda$ be $\bm{p}(\bm{l},\theta)$, $\bm{l} = (x,y)$, with respect to the
  reference frame of map $\bm{M}$. Let the range sensor measure range scan
  $\mathcal{S}_R$. The objective is the estimation of $\bm{p}$ given
  $\mathcal{S}_R$, $\bm{M}$, and $\lambda$.
\end{customprb}

\begin{figure}\vspace{0.4em}
  \input{./figures/face_top.tex}
  \vspace{-1.75cm}
  \caption{\small Top: the map of an environment and the pose of a panoramic 2D
           LIDAR sensor (magenta). Given a LIDAR sensor's 2D measurement, at its
           core, CBGL disperses pose hypotheses within the map and ranks them
           ascendingly (bottom) according to the value of the Cumulative
           Absolute Error per Ray metric (Eq. (\ref{eq:caer})). This
           ranking may estimate the pose of the sensor (a) quickly due to the
           metric's low computational complexity, and (b) accurately due to (i)
           proportionality between the pose estimate error and the value of the
           metric for pose estimates in a neighbourhood of the sensor's pose,
           and (ii) lack of disproportionality outside of that neighbourhood
           (Fig. \ref{fig:motivation_caer})
           }
  \label{fig:face}
\end{figure}

For the solution to Problem \ref{prob:the_problem} this article introduces
CBGL: a single-shot Monte Carlo method
(a) which makes no assumptions regarding the structure or the particulars of the
sensor's environment or the sensor's characteristics,
(b) whose pose errors exhibit robustness to varying sensor angular range, and
(c) which operates with three optionally-set and intuitive parameters, which
trade accuracy for execution time.
The central contributions of the article are:
\begin{itemize}
  \item To the best of the author's knowledge the fastest Monte Carlo global
        localisation method that employs a 2D LIDAR that achieves higher
        pose discovery rates than state-of-the-art methods
  \item The extension and validation of the Cumulative Absolute Error per Ray
        metric's ability to estimate pose error hierarchies solely from real
        and virtual range scans, extended from scan--to--map-scan matching
        (\texttt{sm2}) during pose-tracking, where pose estimates are ``few"
        ($\leq$$2^5$) and their errors are ``small", to the context of global
        localisation, where the set of hypotheses and their errors may be
        arbitrarily large
  \item The thorough evaluation of the proposed method against (a) established
        state-of-the-art localisation methods, (b) real and publicly available
        benchmark conditions, and (c) varying characteristics of environments
        and sensors, which target real conditions, that pose hindrances to
        global localisation methods
\end{itemize}

The rest of the article is structured as follows: Section
\ref{section:definitions} provides necessary definitions and the notation
employed. Section \ref{section:sota} gives a brief review of
solutions to problem \ref{prob:the_problem}, and the relation of the proposed
method to them. The latter's methodology is presented in section
\ref{section:the_proposed_method}, its evaluation in section
\ref{section:results}, and its limitations in section
\ref{section:characterisation}. Section \ref{section:finale} concludes this
study.

\section{Definitions and Notation}
  \label{section:definitions}
  Let $\mathcal{A} = \{\alpha_i: \alpha_i \in \mathbb{R}\}$,
$i \in \texttt{I} = \langle 0,1,\dots,n-1 \rangle$, denote a set of $n$ elements,
$\langle\cdot\rangle$ denote an ordered set, $\mathcal{A}_{\uparrow}$ the
set $\mathcal{A}$ ordered in ascending order, the bracket notation
$\mathcal{A}[\texttt{I}] = \mathcal{A}$ denote indexing, and notation
$\mathcal{A}_{k:l}$, $0 \leq k \leq l$, denote limited indexing:
$\mathcal{A}_{k:l}= \{\mathcal{A}[k], \mathcal{A}[k+1], \dots, \mathcal{A}[l]\}$.

\begin{definition}
  \label{def:definition_0} \textit{Pose and pose estimate}.---The object of
  localisation is the estimation of the---fundamentally unknown---3DoF pose
  $\bm{p}$ of a sensor or robot: $\hat{\bm{p}} = (\hat{\bm{l}}, \hat{\theta})$,
  where $\hat{\bm{l}} = (\hat{x},\hat{y}) \in \mathbb{R}^2$ is its location on
  the 2D plane and $\hat{\theta} \in [-\pi, \pi)$ rad its orientation relative
  to the positive $x$ axis.
\end{definition}

\begin{definition}
  \label{def:definition_1} \textit{Range scan captured from a 2D LIDAR
  sensor}.---A range scan $\mathcal{S}$, measured by a 2D LIDAR sensor,
  consisting of $N_s$ rays over an angular range $\lambda \in (0,2\pi]$, is an
  ordered map of angles to distances between objects and the sensor within its
  radial range $r_{\max}$: $\mathcal{S} : \Theta \rightarrow \mathbb{R}_{\geq
  0}$, $\Theta = \{\theta_n \in [-\frac{\lambda}{2}, +\frac{\lambda}{2}) :
  \theta_n = -\frac{\lambda}{2} + \lambda \frac{n}{N_s}$, $n = 0,1,\dots,
  N_s$$-$$1$$\}$.  Angles $\theta_n$ are expressed relative to the sensor's
  heading, in the sensor's frame of reference \cite{Cooper2018c}.
\end{definition}

\begin{definition}
  \label{def:definition_2} \textit{Map-scan}.---A map-scan is a virtual scan
  that encapsulates the same pieces of information as a scan derived from a
  physical sensor (Def. \ref{def:definition_1}). Contrary to the latter it
  refers to distances within a map $\bm{M}$ of an environment rather than within
  the environment itself; the latter is represented by $\bm{M}$ usually in
  Occupancy Grid or ordered Point Cloud form. A map-scan
  $\mathcal{S}_V^{\bm{M}}(\hat{\bm{p}})$ is derived by locating intersections
  of rays emanating from a sensor's pose estimate $\hat{\bm{p}}$ and the
  boundaries of $\bm{M}$.
\end{definition}

\begin{definition}
  \label{def:definition_3} \textit{The Cumulative Absolute Error per Ray metric}.---Let
  $\mathcal{S}_p$ and $\mathcal{S}_q$ be two range scans (Defs.
  \ref{def:definition_1}, \ref{def:definition_2}), equal in angular range
  $\lambda$ and size $N_s$. The value of the Cumulative Absolute Error per Ray
  (CAER) metric $\psi \in \mathbb{R}_{\geq 0}$ between $\mathcal{S}_p$ and
  $\mathcal{S}_q$ is given by
  \begin{align}
    \psi(\mathcal{S}_p,\mathcal{S}_q) \triangleq \sum\limits_{n=0}^{N_s-1} \Big| \mathcal{S}_p[\theta_n]-\mathcal{S}_q[\theta_n]\Big| \label{eq:caer}
  \end{align}
\end{definition}

\begin{definition}
  \label{def:definition_6} \textit{Pose densities}.---The locational and
  angular density, $d_{\bm{l}}, d_{\alpha} \in \mathbb{N}$, of a pose set
  $\mathcal{H} = \{\hat{\bm{p}}_i\}$ respectively denote the mean number of
  pose estimates dispersed on a given map per unit area of space and
  per angular cycle.
\end{definition}

\begin{definition}
  \label{def:definition_7} \textit{Admissibility of solution}.---A pose estimate
  $\hat{\bm{p}}(\hat{\bm{l}},\hat{\theta}) \in \mathbb{R}^2 \times [-\pi,+\pi)$,
  may be deemed an admissible solution to Problem \ref{prob:the_problem} iff
  $\|\bm{p}-\hat{\bm{p}}\|_2 \leq \delta_0$ and
  $\|\bm{l}-\hat{\bm{l}}\|_2 \leq \delta_{\bm{l}}$ and
  $|\theta-\hat{\theta}| \leq \delta_\theta$, where
  $\delta_{\bm{l}}, \delta_\theta, \delta_0 \in \mathbb{R}_{> 0}$:
  $\delta_{\bm{l}}^2 + \delta_{\theta}^2 = \delta_0^2$.
\end{definition}

\section{Related Work}
  \label{section:sota}
  The literature concerning the solution to the problem of global localisation
with the use of a 2D LIDAR sensor is rich. A recent and comprehensive survey on
global localisation may be found in \cite{gl_survey_cn}, while a brief overview
is given below.

In broad terms global localisation approaches may be divided into two
categories: (a) approaches that operate in feature space, that is, methods that
extract features from measurements and the map and establish correspondences
between them, and (b) approaches that directly exploit only raw measurements.
In the latter category a number of methods solve the problem in an iterative
Bayesian Monte Carlo fashion, i.e. by dispersing hypotheses within the map and
updating the belief of the robot's pose by incorporating new measurements as it
moves until estimate convergence
\cite{mcl,Wang2018d,Yilmaz2019a,gmcl,Chen2021a}.
Although motion may be assistive---or even required---in certain conditions,
such as those of repetitive map structures, the requirement of motion (a) may
give rise to safety concerns (the robot may not even be visible), and (b)
increases estimation time. The Monte Carlo method proposed in this article
operates directly in measurement space as well but, in contrast, is not
iterative, and does not require motion or more than one measurement. As a
result CBGL is a single-shot global localisation approach that is able to
process more hypotheses in less time, resulting in an improvement in the number
of correctly estimated locations, and making it able to compete against
(traditionally faster) feature-based approaches in terms of execution time.

Research on the feature-based approaches has been more extensive due to the
richness, adaptability, and efficacy of methods originated in the computer
vision field, and their low execution time. Relevant methods mainly perform
detection of key-points in a measurement, followed by the calculation of a
distinctive signature, which is then matched to similarly- and pre-computed
place-signatures
\cite{Kallasi2016a,als_eth,Usman2019,Wang2021b,Meng2021,Hendrikx2021,An2022,Nielsen2023}.
In principle, however, unstructured environments cannot be relied upon for the
existence of features due to their absence or their sparse and fortuitous
distribution (although Deep Neural Network approaches have demonstrated
increased performance in place recognition with the use of 3D LIDARs
\cite{Xu2021,Yin2022,Komorowski2022}). Structured environments, on the other
hand, manifest different features depending on the particularities of the
environment, where features may be present but not in a sufficiently
undisturbed state due to sensor noise or map-to-environment mismatch.
Furthermore, feature-based methods require the tuning of parameters in a
per-environment basis, which hinders the range and degree of their
applicability and efficacy.  In contrast CBGL (a) makes no assumptions on- and
exploits no environmental structure, and (b) uses three parameters, whose
setting is optional and intuitive.

In sum the proposed method retains most of the positive qualities of the two
main approaches to global localisation and avoids their pitfalls: it uses
multiple hypotheses for robustness against uncertainty, assumes no motion,
environmental structure, or parameter tuning in a per-environment or sensor
basis, and is robust against noise.  The motivation of CBGL originates from
seeking to achieve a greater degree of universality, reliability, and
portability across multiple and disparate environments, by aiming to economise
on the use of resources (environmental assumptions; number, type, and cost of
sensors; number of measurements; time). The proposed method is most akin to the
two tested methods in \cite{Filotheou2022g}: all three are single-shot 2D
LIDAR-based Monte Carlo approaches, but CBGL computes a measure of the
alignment of the measurement to the virtual scan captured from each hypothesis
\textit{before} scan--to--map-scan matching, which occurs subsequently and only
for a small subset of the virtual scans most aligned to the
measurement. This, along with the low computational complexity of the alignment
measure that CBGL utilises (Def. \ref{def:definition_3}), greatly diminish its
execution time. So far this alignment measure has only been tested against the
context of \texttt{sm2} in pose tracking, i.e. for pose hypotheses in a
neighbourhood of the sensor's true pose \cite{Filotheou2022f,Filotheou2023a}.

\section{Methodology}
  \label{section:the_proposed_method}
\subsection{Motivation}
\label{subsec:motivation}

The proposed method's motivation lies in the simple but powerful facts that are
in general exhibited through fig. \ref{fig:motivation_caer}: Let a pose
estimate of a 2D LIDAR sensor be within the given map (Def.
\ref{def:definition_0}); then the value of the CAER metric (Def.
\ref{def:definition_3}) between the scan measured by the sensor (Def.
\ref{def:definition_1}) and the map-scan captured from the estimate (Def.
\ref{def:definition_2}) is (a) proportional to both the estimate's location
error and orientation error in a neighbourhood of the sensor's true pose, (b)
not inversely proportional to them outside of it, and (c) greater in value
outside than inside of it. In other words comparing any two estimates of
the sensor's pose in terms of the CAER metric, where at least one is in a
neighbourhood of it, is enough to establish a pose error hierarchy between
them.  This is significant because: while the error of a single pose estimate
is unknowable, the top of a hierarchy of dense-enough pose estimates (Def.
\ref{def:definition_6}) ordered by CAER values may provide the neighbourhood of
the sensor's pose, and hence an admissible estimate of the pose itself (Def.
\ref{def:definition_7}).  Moreover, the CAER metric is calculable from
the assumptions of Problem \ref{prob:the_problem}, with low computational
complexity $\mathcal{O}(N_s)$.  The relationships of proportionality between
the CAER metric of a pose estimate and its location and orientation errors have
been discovered and successfully exploited in the context of non-global
localisation, and specifically in pose-tracking, for the production of lidar
odometry \cite{Filotheou2022f} and the reduction of localisation's pose
estimate error \cite{Filotheou2023a}. In that context estimate errors are close
to the origin, in contrast to the distribution of hypotheses' errors in global
localisation methods.  Fig. \ref{fig:h_and_h_not_fig} (top) shows the relations
between the total errors of hypotheses and their (a) CAER values (left) and (b)
ranks when ordered by CAER ascending (right), which resulted from an
experimental procedure similar to that which produced fig.
\ref{fig:motivation_caer}. The above motivate investigation on whether the CAER
metric could prove equally beneficial in settings more uncertain than
pose-tracking, that is, where the location and orientation of pose estimates
extend farther away from the sensor's true location and orientation.

\begin{figure}\vspace{-1.5cm}
  \subfloat{\hspace{0.5cm}\input{./figures/caer_all.tex}}\vspace{-1.5cm}\\
  \subfloat{\hspace{-0.3cm}\input{./figures/face_bottom.tex}}
  \caption{\small Top: without loss of generality, a typical plot of the
           Cumulative Absolute Error per Ray metric (Eq. (\ref{eq:caer})) of
           $10^6$ hypotheses disperesed in the map of environment WAREHOUSE
           (Fig. \ref{fig:face} and section \ref{subsec:exp_b}). Bottom:
           focused view on hypotheses with location error close to the origin}
  \vspace{-0.5cm}
  \label{fig:motivation_caer}
\end{figure}

\subsection{The CBGL Algorithm}

In order to test the efficacy of the CAER metric in aiding the solution of
Problem \ref{prob:the_problem}, we introduce the CAER-Based Global Localisation
algorithm (CBGL), described in block diagram form in fig.
\ref{fig:block_system} and in pseudocode in Algorithm
\ref{alg:cbgl}.\footnote{For a rigorous mathematical formulation of the
hypothesis underpinning CBGL see \cite{cbglarxiv}.} Given map $\bm{M}$, CBGL
first generates a set of pose hypotheses $\mathcal{H}$ (Def.
\ref{def:definition_0}). Their positions are randomly generated uniformly
within the map's traversable space; their orientations within $[-\pi,\pi)$ rad.
Then from these poses CBGL computes map-scans (Def.  \ref{def:definition_2}).
Given these map-scans and a LIDAR's 2D measurement $\mathcal{S}_R$ (Def.
\ref{def:definition_1}), CBGL subsequently computes the CAER value (Def.
\ref{def:definition_3}) associated with each pose hypothesis.  It then ranks
them in ascending order and selects the $k$ estimates with the least CAER
values in an attempt to estimate the $k$ hypotheses with the least pose error,
producing set $\mathcal{H}_1$ (Alg. \ref{alg:bottom_k}). Estimation in this
sense rests on the motivation of subsection \ref{subsec:motivation}. Fig.
\ref{fig:face} (bottom) indicatively depicts such a pose hierarchy.

One challenge is choosing such $k$, $d_{\bm{l}}$, and $d_\alpha$ (Def.
\ref{def:definition_6}) that, given pose estimate error requirements
$\delta_{\bm{l}}$, $\delta_{\theta}$, CBGL produces admissible pose estimates
(Def. \ref{def:definition_7}) while being executed in timely manner.  Given the
method's Monte Carlo nature, optimistically, the only option for increasing the
accuracy of the final pose estimate by a factor of two would be to double
locational density $d_{\bm{l}}$. Instead of doing that---and thereby doubling
the method's execution time---subsequent to the calculation of set
$\mathcal{H}_1$, CBGL scan--to--map-scan matches the map-scans captured from
the pose estimates of $\mathcal{H}_1$ against the range scan measured by the
real sensor (\texttt{sm2}) \cite{Vasiljevic2016c,Filotheou2023a}, producing
pose set $\mathcal{H}_2$ (Alg. \ref{alg:sm2}).

Matching allows for (a) the correction of the pose of true positive estimates,
(b) by the same token the potential divergence of spurious, false positive
pose estimates, and hence their elimination as pose estimate candidates, (c)
the decoupling of the final pose estimate's error from the method's chosen
densities, and (d) the production of finer pose estimates without excessive
increase in execution time. The principle of \texttt{sm2} is depicted in fig.
\ref{fig:sm2_evolution}.

CBGL's output is the pose estimate with the least CAER value within the
group of $k$ matched estimates.

\begin{figure}[H]\vspace{1.3cm}
  \input{./figures/characterisation/sm2_evolution.tex}
  \caption{\small Depiction of the process of scan--to--map-scan matching
           (\texttt{sm2}) in principle, which takes place for all
           $\hat{\bm{p}}_i = \mathcal{H}_1[i]$ (Alg.
           \ref{alg:cbgl}, line \ref{alg:cbgl:h2}), $i=0,1,\dots,k$$-$$1$. The registration of map-scan
           $\mathcal{S}_V^{\bm{M}}(\hat{\bm{p}}[s])$ to $\mathcal{S}_R$ at step
           $s$ results in pose $\hat{\bm{p}}[s$$+$$1]$, where
           $\hat{\bm{p}}_i[0] = \mathcal{H}_1[i]$. The error of
           $\hat{\bm{p}}[s$$+$$1]$ is, in principle, reduced compared to
           $\hat{\bm{p}}[s]$, and the next map-scan is captured from it. The
           process is iterative and completes upon estimate convergence}
  \label{fig:sm2_evolution}
\end{figure}

\begin{figure}[H]
  \subfloat{\label{fig:cbgl}     \tikzset{every picture/.style={line width=0.75pt}} 

\hspace{-0.3cm}
\begin{tikzpicture}[x=0.75pt,y=0.75pt,yscale=-0.5,xscale=0.5]

\draw  [dash pattern={on 0.84pt off 2.51pt}] (132,113) -- (386.5,113) -- (386.5,329) -- (132,329) -- cycle ;
\draw    (178.5,106) -- (178.5,183) ;
\draw [shift={(178.5,183)}, rotate = 270] [color={rgb, 255:red, 0; green, 0; blue, 0 }  ][line width=0.75]    (10.93,-3.29) .. controls (6.95,-1.4) and (3.31,-0.3) .. (0,0) .. controls (3.31,0.3) and (6.95,1.4) .. (10.93,3.29)   ;
\draw    (299.5,158) -- (299.5,183) ;
\draw [shift={(299.5,183)}, rotate = 270] [color={rgb, 255:red, 0; green, 0; blue, 0 }  ][line width=0.75]    (10.93,-3.29) .. controls (6.95,-1.4) and (3.31,-0.3) .. (0,0) .. controls (3.31,0.3) and (6.95,1.4) .. (10.93,3.29)   ;
\draw    (219.5,105.33) -- (219.5,183) ;
\draw [shift={(219.5,183)}, rotate = 270] [color={rgb, 255:red, 0; green, 0; blue, 0 }  ][line width=0.75]    (10.93,-3.29) .. controls (6.95,-1.4) and (3.31,-0.3) .. (0,0) .. controls (3.31,0.3) and (6.95,1.4) .. (10.93,3.29)   ;
\draw    (313.5,104) -- (313.5,125) ;
\draw [shift={(313.5,128)}, rotate = 270] [color={rgb, 255:red, 0; green, 0; blue, 0 }  ][line width=0.75]    (10.93,-3.29) .. controls (6.95,-1.4) and (3.31,-0.3) .. (0,0) .. controls (3.31,0.3) and (6.95,1.4) .. (10.93,3.29)   ;

\draw    (230.79,94.31) -- (286,94.31) -- (286,125) ;
\draw [shift={(286,128)}, rotate = 270] [color={rgb, 255:red, 0; green, 0; blue, 0 }  ][line width=0.75]    (10.93,-3.29) .. controls (6.95,-1.4) and (3.31,-0.3) .. (0,0) .. controls (3.31,0.3) and (6.95,1.4) .. (10.93,3.29)   ;

\draw    (160.79,94.31) -- (125,94.31) -- (125,257) -- (217.67,257) ;
\draw [shift={(219.67,257)}, rotate = 180] [color={rgb, 255:red, 0; green, 0; blue, 0 }  ][line width=0.75]    (10.93,-3.29) .. controls (6.95,-1.4) and (3.31,-0.3) .. (0,0) .. controls (3.31,0.3) and (6.95,1.4) .. (10.93,3.29)   ;
\draw    (220.33,83.17) -- (220.33,67) -- (125,67) -- (125,94.31) ;
\draw    (240.5,215) -- (240.5,240) ;
\draw [shift={(240.5,242)}, rotate = 270] [color={rgb, 255:red, 0; green, 0; blue, 0 }  ][line width=0.75]    (10.93,-3.29) .. controls (6.95,-1.4) and (3.31,-0.3) .. (0,0) .. controls (3.31,0.3) and (6.95,1.4) .. (10.93,3.29)   ;
\draw    (395,202) -- (356.85,202) ;
\draw [shift={(354.85,202.01)}] [color={rgb, 255:red, 0; green, 0; blue, 0 }  ][line width=0.75]    (10.93,-3.29) .. controls (6.95,-1.4) and (3.31,-0.3) .. (0,0) .. controls (3.31,0.3) and (6.95,1.4) .. (10.93,3.29)   ;
\draw    (240.5,269) -- (240.5,294) ;
\draw [shift={(240.5,296)}, rotate = 270] [color={rgb, 255:red, 0; green, 0; blue, 0 }  ][line width=0.75]    (10.93,-3.29) .. controls (6.95,-1.4) and (3.31,-0.3) .. (0,0) .. controls (3.31,0.3) and (6.95,1.4) .. (10.93,3.29)   ;
\draw    (395,310) -- (352.85,310) ;
\draw [shift={(350.85,310)}] [color={rgb, 255:red, 0; green, 0; blue, 0 }  ][line width=0.75]    (10.93,-3.29) .. controls (6.95,-1.4) and (3.31,-0.3) .. (0,0) .. controls (3.31,0.3) and (6.95,1.4) .. (10.93,3.29)   ;
\draw    (125,257) -- (125,310) -- (145.67,310) ;
\draw [shift={(145.67,310)}, rotate = 180] [color={rgb, 255:red, 0; green, 0; blue, 0 }  ][line width=0.75]    (10.93,-3.29) .. controls (6.95,-1.4) and (3.31,-0.3) .. (0,0) .. controls (3.31,0.3) and (6.95,1.4) .. (10.93,3.29)   ;
\draw    (240.5,322) -- (240.5,347) ;
\draw [shift={(240.5,349)}, rotate = 270] [color={rgb, 255:red, 0; green, 0; blue, 0 }  ][line width=0.75]    (10.93,-3.29) .. controls (6.95,-1.4) and (3.31,-0.3) .. (0,0) .. controls (3.31,0.3) and (6.95,1.4) .. (10.93,3.29)   ;

\draw (179,95) node   [align=left] {\tiny $\mathcal{S}_R$};
\draw    (227,133.5) -- (381,133.5) -- (381,158.5) -- (227,158.5) -- cycle  ;
\draw (304,147) node   [align=left] {\tiny Generate Hypotheses};
\draw (318,172) node   [align=left] {\tiny $\mathcal{H}$};
\draw (220.33,94.83) node   [align=left] {\tiny $\bm{M}$};
\draw (318.33,84.83) node   [align=left] {\tiny $(d_{\bm{l}}, d_{\alpha})$};
\draw    (223,244.5) -- (259,244.5) -- (259,269.5) -- (223,269.5) -- cycle  ;
\draw (241,257) node   [align=left] {\tiny \texttt{sm2}};
\draw  [fill={rgb, 255:red, 255; green, 254; blue, 229 }  ,fill opacity=1 ]  (146.5,189.5) -- (347.5,189.5) -- (347.5,214.5) -- (146.5,214.5) -- cycle  ;
\draw (247,202) node   [align=left] {\tiny Estimate Bottom-$k$ of Pose Errors};
\draw (397,194.71) node [anchor=north west][inner sep=0.75pt]   [align=left] {\tiny $k$$\ll$$|\mathcal{H}|$};
\draw  [fill={rgb, 255:red, 255; green, 254; blue, 229 }  ,fill opacity=1 ]  (146.5,297.5) -- (347.5,297.5) -- (347.5,322.5) -- (146.5,322.5) -- cycle  ;
\draw (247,310) node   [align=left] {\tiny Estimate Bottom-$k$ of Pose Errors};
\draw (397,302.71) node [anchor=north west][inner sep=0.75pt]   [align=left] {\tiny $k$$=$$1$};
\draw (280,232) node   [align=left] {\tiny $\mathcal{H}_1 \subseteq \mathcal{H}$};
\draw (324,284) node   [align=left] {\tiny $\mathcal{H}_2 \not\subseteq \mathcal{H}$; $\mathcal{H}_2 \neq \mathcal{H}_1$};
\draw (241,366) node   [align=left] {\tiny $\hat{\bm{p}}$};

\end{tikzpicture}}
  \subfloat{\label{fig:bottom_k} \tikzset{every picture/.style={line width=0.75pt}} 

\hspace{-0.9cm}
\begin{tikzpicture}[x=0.75pt,y=0.75pt,yscale=-0.5,xscale=0.5]

\draw  [draw opacity=0][fill={rgb, 255:red, 255; green, 254; blue, 229 }  ,fill opacity=1 ] (149.5,52) -- (400.5,52) -- (400.5,336) -- (149.5,336) -- cycle ;
\draw  [draw opacity=0][fill={rgb, 255:red, 217; green, 240; blue, 163 }  ,fill opacity=1 ][dash pattern={on 4.5pt off 4.5pt}] (168,76) -- (380.5,76) -- (380.5,270) -- (168,270) -- cycle ;
\draw  [draw opacity=0][fill={rgb, 255:red, 173; green, 221; blue, 142 }  ,fill opacity=1 ][dash pattern={on 0.84pt off 2.51pt}] (190.5,96) -- (360.5,96) -- (360.5,210) -- (190.5,210) -- cycle ;
\draw    (135.5,37.83) -- (135.5,185.83) -- (212,185.83) ;
\draw [shift={(214,185.83)}, rotate = 180.12] [color={rgb, 255:red, 0; green, 0; blue, 0 }  ][line width=0.75]    (10.93,-3.29) .. controls (6.95,-1.4) and (3.31,-0.3) .. (0,0) .. controls (3.31,0.3) and (6.95,1.4) .. (10.93,3.29)   ;
\draw    (225.5,36.83) -- (225.5,114.83) [line width=0.5];
\draw [shift={(225.5,116.83)}, rotate = 270] [color={rgb, 255:red, 0; green, 0; blue, 0 }  ][line width=0.75]    (10.93,-3.29) .. controls (6.95,-1.4) and (3.31,-0.3) .. (0,0) .. controls (3.31,0.3) and (6.95,1.4) .. (10.93,3.29)   ;
\draw    (275.5,143.83) -- (275.5,168.83) ;
\draw [shift={(275.5,170.83)}, rotate = 270] [color={rgb, 255:red, 0; green, 0; blue, 0 }  ][line width=0.75]    (10.93,-3.29) .. controls (6.95,-1.4) and (3.31,-0.3) .. (0,0) .. controls (3.31,0.3) and (6.95,1.4) .. (10.93,3.29)   ;
\draw    (275.5,326.83) -- (275.5,361.83) ;
\draw [shift={(275.5,363.83)}, rotate = 270] [color={rgb, 255:red, 0; green, 0; blue, 0 }  ][line width=0.75]    (10.93,-3.29) .. controls (6.95,-1.4) and (3.31,-0.3) .. (0,0) .. controls (3.31,0.3) and (6.95,1.4) .. (10.93,3.29)   ;
\draw    (414.5,38.83) -- (414.5,131.83) -- (345.5,131.83) ;
\draw [shift={(343.5,131.83)}, rotate = 360] [color={rgb, 255:red, 0; green, 0; blue, 0 }  ][line width=0.75]    (10.93,-3.29) .. controls (6.95,-1.4) and (3.31,-0.3) .. (0,0) .. controls (3.31,0.3) and (6.95,1.4) .. (10.93,3.29)   ;
\draw    (414.5,113.83) -- (414.5,315.83) -- (325.5,315.83) ;
\draw [shift={(323.5,315.83)}, rotate = 360] [color={rgb, 255:red, 0; green, 0; blue, 0 }  ][line width=0.75]    (10.93,-3.29) .. controls (6.95,-1.4) and (3.31,-0.3) .. (0,0) .. controls (3.31,0.3) and (6.95,1.4) .. (10.93,3.29)   ;
\draw    (275.5,197.83) -- (275.5,232.83) ;
\draw [shift={(275.5,234.83)}, rotate = 270] [color={rgb, 255:red, 0; green, 0; blue, 0 }  ][line width=0.75]    (10.93,-3.29) .. controls (6.95,-1.4) and (3.31,-0.3) .. (0,0) .. controls (3.31,0.3) and (6.95,1.4) .. (10.93,3.29)   ;
\draw    (136.52,314) -- (228.67,314) ;
\draw [shift={(230.67,314)}, rotate = 180.19] [color={rgb, 255:red, 0; green, 0; blue, 0 }  ][line width=0.75]    (10.93,-3.29) .. controls (6.95,-1.4) and (3.31,-0.3) .. (0,0) .. controls (3.31,0.3) and (6.95,1.4) .. (10.93,3.29)   ;
\draw    (275.5,261.83) -- (275.5,295.83) ;
\draw [shift={(275.5,297.83)}, rotate = 270] [color={rgb, 255:red, 0; green, 0; blue, 0 }  ][line width=0.75]    (10.93,-3.29) .. controls (6.95,-1.4) and (3.31,-0.3) .. (0,0) .. controls (3.31,0.3) and (6.95,1.4) .. (10.93,3.29)   ;

\draw (225,54) node [anchor=north west][inner sep=0.75pt]   [align=left] {\tiny Estimate bottom-$k$ of pose errors};
\draw (285,77) node [anchor=north west][inner sep=0.75pt]   [align=left] {\tiny compute ranks};
\draw (264,97) node [anchor=north west][inner sep=0.75pt]   [align=left] {\tiny compute CAERs};
\draw (137,27) node     [align=left] {\tiny $\mathcal{S}_R$};
\draw (226.5,27) node   [align=left] {\tiny $\bm{M}$};
\draw (416,27) node  [align=left] {\tiny $\mathcal{P}$};
\draw    (212.5,119.33) -- (338.5,119.33) -- (338.5,144.33) -- (212.5,144.33) -- cycle  ;
\draw (275.5,133) node   [align=left] {\tiny  \ \ \ \ \ \texttt{scan\_map} \ \ \ \ \ };
\draw (303,157.83) node   [align=left] {\tiny $\{\mathcal{S}_V\}$};
\draw    (219.5,172.67) -- (331.5,172.67) -- (331.5,197.67) -- (219.5,197.67) -- cycle  ;
\draw (275.5,185.17) node   [align=left] {\tiny  \ \ \ \ \ \ \ $\psi$ \ \ \ \ \ \ };
\draw    (224.5,237) -- (324.5,237) -- (324.5,262) -- (224.5,262) -- cycle  ;
\draw (274.5,249.5) node   [align=left] {\tiny $\texttt{sort}(\Psi,\texttt{asc})$};
\draw (288,224) node   [align=left] {\tiny $\Psi$};
\draw    (233.25,302) -- (318.25,302) -- (318.25,327) -- (233.25,327) -- cycle  ;
\draw (241,285) node   [align=left] {\tiny $\texttt{I}^{\ast}, \Psi_{\uparrow}$: };
\draw (330,285) node   [align=left] {\tiny $\Psi[\texttt{I}^{\ast}] = \Psi_{\uparrow}$};

\draw (276.75,315) node   [align=left] {\tiny  $\mathcal{P}[\texttt{I}^{\ast}]_{1:k} $};
\draw (121,307) node [anchor=north west][inner sep=0.75pt]   [align=left] {\tiny $k$};
\draw (275.5,376.5) node   [align=left] {\tiny $\mathcal{P}_{\triangledown} = \{\mathcal{P}[\texttt{I}^{\ast}[0]], \mathcal{P}[\texttt{I}^{\ast}[1]], \dots, \mathcal{P}[\texttt{I}^{\ast}[k$$-$$1]]\}$};

\end{tikzpicture}}
  \caption{\small CBGL in block diagram form. Left: Given map $\bm{M}$, CBGL
           first generates a set of pose hypotheses $\mathcal{H}$. Then it
           estimates the $k$ hypotheses with the least pose error (right; Alg.
           \ref{alg:bottom_k}).
           As a final step, it scan--to--map-scan
           matches these to $\mathcal{S}_R$ for finer estimation
           (Alg. \ref{alg:sm2}).
           CBGL's output pose estimate is that with
           the minimum CAER among the $k$ matched estimates. See Algs.
           \ref{alg:cbgl}, \ref{alg:bottom_k}, \ref{alg:sm2} for notation
           }
  \label{fig:block_system}
\end{figure}

\begin{algorithm}[H]
  \caption{\texttt{CBGL}}
  \begin{spacing}{1.0}
  \begin{algorithmic}[1]
    \REQUIRE $\mathcal{S}_R$, $\lambda$, $\bm{M}$, $(d_{\bm{l}}, d_\alpha)$, $k$
    \ENSURE Pose estimate of sensor measuring range scan $\mathcal{S}_R$ 
    \STATE $A \leftarrow \texttt{calculate\_area}(\texttt{free}(\bm{M}))$
    \STATE $\mathcal{H}, \mathcal{H}_1, \mathcal{H}_2 \leftarrow \{\varnothing\}$
    \FOR {$i \leftarrow 0,1,\dots,d_{\bm{l}} \cdot A-1$}
      \STATE \small $(\hat{x},\hat{y},\hat{\theta}) \leftarrow \texttt{rand()}$: $(x,y) \in \texttt{free}(\bm{M})$, $\hat{\theta} \in [-\pi,+\pi)$
      \FOR {$j \leftarrow 0,1,\dots, d_{\alpha}-1$}
        \STATE $\mathcal{H} \leftarrow \{\mathcal{H}, (\hat{x}, \hat{y}, \hat{\theta} + j \cdot 2\pi / d_{\alpha})\}$     \label{alg:cbgl:h}
      \ENDFOR
    \ENDFOR
    \STATE $\mathcal{H}_1 \leftarrow$ \texttt{bottom}$\_k\_\texttt{poses}(\mathcal{S}_R, \bm{M}, \mathcal{H}, k, \lambda)$ \hfill {\small (Alg. \ref{alg:bottom_k}}) \label{alg:cbgl:h1}
    \FOR {$i \leftarrow 0,1,\dots,|\mathcal{H}_1|-1$}
      \STATE \footnotesize $\mathcal{H}_2 \leftarrow \{\mathcal{H}_2, \texttt{sm2}(\mathcal{S}_R, \lambda, \bm{M}, \mathcal{H}_1[i])\}$ \hfill {(Alg. \ref{alg:sm2}---e.g. \texttt{x1} \cite{Filotheou2023a})}\label{alg:cbgl:h2}
    \ENDFOR
    \RETURN \texttt{bottom}$\_k\_\texttt{poses}(\mathcal{S}_R, \bm{M}, \mathcal{H}_2, 1, \lambda)$
  \end{algorithmic}
  \end{spacing}
  \label{alg:cbgl}
\end{algorithm}

\begin{algorithm}[H]
  \caption{\texttt{bottom}\_$k$\_\texttt{poses}}
  \begin{spacing}{1.0}
  \begin{algorithmic}[1]
    \REQUIRE $\mathcal{S}_R$, $\bm{M}$, $\mathcal{P}$, $k$, $\lambda$
    \ENSURE Set of $k$ poses of $\mathcal{P}$ with least CAER values, $\mathcal{P}_{\triangledown}$
    \STATE $\Psi \leftarrow \{\varnothing \}$
    \FOR {$q \leftarrow 0,1,\dots,|\mathcal{P}|-1$}
      \STATE $\mathcal{S}_V^{\hspace{1pt} q} \leftarrow \mathcal{S}_V^{\bm{M}}(\mathcal{P}[q]) = \texttt{scan\_map}(\bm{M}, \mathcal{P}[q], \lambda)$
      \STATE $\Psi \leftarrow \{\Psi, \texttt{CAER}(\mathcal{S}_R, \mathcal{S}_V^{\hspace{1pt} q}) \}$ \hfill {\small (Eq. (\ref{eq:caer})})
    \ENDFOR
    \STATE $[\Psi_{\uparrow}, \texttt{I}^{\ast}] \leftarrow \texttt{sort}(\Psi, \texttt{asc})$, such that $\Psi[\texttt{I}^{\ast}] = \Psi_{\uparrow}$
    \RETURN $\mathcal{P}_{\triangledown} = \{\mathcal{P}[\texttt{I}^{\ast}[0]], \mathcal{P}[\texttt{I}^{\ast}[1]], \dots, \mathcal{P}[\texttt{I}^{\ast}[k-1]]\}$
  \end{algorithmic}
  \end{spacing}
  \label{alg:bottom_k}
\end{algorithm}

\begin{algorithm}[H]
  \caption{\texttt{sm2}}
  \begin{spacing}{1.0}
  \begin{algorithmic}[1]
    \REQUIRE $\mathcal{S}_R$, $\lambda$, $\bm{M}$, $\hat{\bm{p}}$
    \ENSURE $\hat{\bm{p}}$ $+$ correction that aligns $\mathcal{S}_V^{\bm{M}}(\hat{\bm{p}})$ to $\mathcal{S}_R$
    \STATE $\mathcal{S}_V \leftarrow \texttt{scan\_map}(\bm{M}, \hat{\bm{p}}, \lambda)$
    \STATE $\bm{\Delta p} \leftarrow \texttt{scan-match}(\mathcal{S}_R,\mathcal{S}_V)$ \hfill {\footnotesize (e.g. \texttt{ICP}\cite{Vizzo2023}, \texttt{FSM}\cite{Filotheou2022f}})
    \RETURN $\hat{\bm{p}} + \bm{\Delta p}$
  \end{algorithmic}
  \end{spacing}
  \label{alg:sm2}
\end{algorithm}

\section{Experimental Evaluation}
  \label{section:results}
  This section focuses on evaluating the performance of state of the art methods
in global localization and CBGL in addressing Problem \ref{prob:the_problem} in
static environments, varying environmental
conditions, sensor configurations, and map representations. CBGL's
parameters are set to $(d_{\bm{l}},d_{\alpha},k) = (40, 2^5, 10)$
after initial tests with the real dataset used in subsection
\ref{subsec:exp_a}. The rationale of choosing appropriate
$d_{\bm{l}}$ and $d_{\alpha}$ is depicted in fig. \ref{fig:a:determine_40_32};
$k$ is chosen as such in order to retain a high-enough true positive discovery
rate without significant increase in execution time. The locational threshold
$\delta_{\bm{l}} = 0.5$ m is used as a tighter solution admissibility criterion
than that in \cite{Filotheou2022g}; the thresold itself was
determined through experimental procedure with a YDLIDAR TG30 sensor (footnote
3; ibid).
References to sets $\mathcal{H}_{\ast}$ are made to fig. \ref{fig:block_system}
and lines
\ref{alg:cbgl:h}, \ref{alg:cbgl:h1}, and \ref{alg:cbgl:h2} of
Alg.  \ref{alg:cbgl}.
Tests are performed with a processor of $12$ threads and clock speed $4.00$ GHz.

\subsection{Experiments in real conditions}
\label{subsec:exp_a}

The first type of test is conducted using a Hokuyo UTM-30LX sensor, whose
angular range $\lambda = 3\pi/2$ rad and radial range $r_{\max} = 30.0$ m,
in the  Electrical and Computer Engineering Department's Laboratory of Computer
Systems Architecture (CSAL), of the Aristotle University of Thessaloniki, an
Occupancy Grid map of which is depicted in fig.
\ref{fig:a:map_and_trajectory}. The map's resolution is $0.01$ m / pixel, and
it was constructed via ROS package \texttt{open\_karto}. The sensor was mounted
on a Robotnik RB1 robot, which was teleoperated within the environment while
scans were being recorded.  This resulted in $N_E = 6669$ range scans, whose
number of rays are downsampled by a factor of four before being inputted to
CBGL and Advanced Localization System (ALS) \cite{als_jp}. The latter
implements Free-Space Features \cite{als_eth} and, contrary to CBGL, is not a
single-shot method; however, it is selected for comparison against CBGL due to
the fact that it is the only state of the art global localisation method which
exhibits feasible execution times with respect to the collected range dataset's
volume (see bottom of fig.  \ref{fig:b:inliers_per_pose} for indicative
execution times of state of the art methods). CBGL's internal \texttt{sm2}
method is chosen to be PLICP \cite{Censi2008c} due to its low execution time
and the sensor's non-panoramic field of view. The top of fig.
\ref{fig:a:awesomeness} depicts the proportion of output pose estimates from
each method whose position and orientation error is lower than outlier
thresholds $\delta_{\bm{l}}$, $\delta_{\theta}$; at the bottom they are
depicted exclusively for CBGL's outputs and its internal pose sets. Table
\ref{tbl:a} provides a summary of the errors and execution times of the two
methods.

From the experimental evidence it is clear that (a) CBGL manages to produce
admissible solutions to Problem \ref{prob:the_problem} $991$ times out of a
thousand for an outliers' locational threshold $\delta_{\bm{l}} = 0.5$ m when
an angular threshold $\delta_{\theta}$ is not considered, and (b) CBGL
outperforms ALS in terms of (i) position and orientation errors, (ii) number of
pose estimates within all locational and angular thresholds, and (iii) execution
time.

\begin{table}[H]\footnotesize
\begin{tabular}{@{}lcccccc@{}}
     & \multicolumn{2}{l}{Position Err. {[}m{]}} & \multicolumn{2}{l}{Orientation Err. {[}rad{]}} & \multicolumn{2}{l}{Exec. Time {[}sec{]}} \\
     & Mean              & std              & Mean                & std                 & Mean                & std                \\ \midrule
ALS  & $0.500 $          & $0.265 $         & $1.956 $            & $1.167 $            & $6.15 $             & $5.32 $            \\
CBGL & $0.041 $          & $0.045 $         & $0.011 $            & $0.019 $            & $1.61 $             & $0.06 $            \\ \bottomrule
\end{tabular}
  \caption{\small Mean and standard deviation of (a) errors of ALS and CBGL with
                regard to position and orientation and (b) their execution times,
                with regard to experiments in real conditions}
\label{tbl:a}
\end{table}

\begin{figure}
  \vspace{-0.4cm}
  \input{./figures/experiments/a/determine_40_32.tex}
  \vspace{-0.4cm}
  \caption{\small Top row and right column: histograms of the number of times
           out of $N_E = 6669$ attempts of CBGL at global localisation
           when exactly $n \in [0,k] = [0,10]$ pose estimates belonging to set
           $\mathcal{H}_1$ exhibited pose errors lower than
           $\delta_0 = 0.5$ (m$^2$ + rad$^2$)$^{1/2}$ (e.g. for
           ($\delta_{\bm{l}}, \delta_{\theta}) = (0.3,0.4) [\text{m},
           \text{rad}]$). For densities $(d_{\bm{l}},d_{\alpha}) = (40, 2^5)$
           (top right) this number is strictly increasing with $n$, which would
           be an aspect to be retrospectively expected of a global localisation
           method. Denser configurations adhere to the same pattern as well but
           would require more execution resources. Middle block: the resulting
           percent proportion of pose estimates whose pose error is lower than
           $\delta_0$ for varying pose densities. Bottom block: the resulting
           distribution of corresponding execution times
           }
  \label{fig:a:determine_40_32}
\end{figure}

\begin{figure}
  \vspace{-0.7cm}
  \input{./figures/experiments/a/trajectory.tex}
  \vspace{-0.7cm}
  \caption{\small The map of the real environment CSAL (black), the trajectory
           of the sensor (blue), and CBGL's estimated positions of the sensor
           (green). A total of $N_E = 6669$ pose estimations take place,
           $99.1\%$ of which result in position errors lower than
           $\delta_{\bm{l}} = 0.5$ m. See fig. \ref{fig:a:awesomeness}
           for the respective percentages of position and orientation errors
           for varying admissibility thresholds $\delta_{\bm{l}}$ and
           $\delta_{\theta}$.
           Estimation is performed for each sensor
           pose independently of previous estimates and measurements. Sensor
           poses for which CBGL's output exhibits position error larger than
           $\delta_{\bm{l}} = 0.5$ m are marked with red, sources of great
           range noise with cyan, and regions around doors with purple
           }
  \label{fig:a:map_and_trajectory}
  \vspace{-0.5cm}
\end{figure}

\begin{figure}
  \vspace{-0.3cm}
  \input{./figures/experiments/a/awesomeness.tex}
  \vspace{0.01cm}
  \caption{\small Proportions of pose estimates whose position and orientation
           error is lower than corresponding thresholds $\delta_{\bm{l}}$ and
           $\delta_{\theta}$. Top: ALS vs CBGL. Bottom: CBGL and internal pose
           estimate sets.  Approximately $77\%$ of all bottom-$k$ pose
           estimates---the contents of $\mathcal{H}_1$ sets---exhibit
           position errors lower than $\delta_{\bm{l}} = 0.5$ m for $k=10$, and
           so do over $99\%$ of CBGL's output pose estimates. The improvement in
           position and orientation induced by scan--to--map-scan matching is
           captured by the difference between the output (i.e.
           $\mathcal{H}_2[\arg \min \psi(\mathcal{S}_R, \mathcal{S}_V^{\bm{M}}(\mathcal{H}_2)]$) and
           $\mathcal{H}_1[\arg \min \psi(\mathcal{S}_R, \mathcal{S}_V^{\bm{M}}(\mathcal{H}_2)]$}
  \vspace{-0.7cm}
  \label{fig:a:awesomeness}
\end{figure}

\subsection{Simulations against sources of uncertainty}
\label{subsec:exp_b}

The second type of test concerns the main limiting factor of global
localisation methods, i.e. uncertainty---: arising e.g. from spurious
measurements, repeatability of surroundings, missing or corrupted range
information, or their combinations. For this reason the experimental procedure
of \cite{Filotheou2022g} is extended here for the two methods tested therein,
i.e. Passive Global Localisation-Fourier-Mellin Invariant matching with
Centroids for translation (PGL-FMIC) and- Point-to-Line ICP (PGL-PLICP), and
then for ALS, Monte Carlo Localisation (MCL) \cite{mcl}, and General Monte
Carlo Localisation (GMCL) \cite{gmcl}. All methods are tested against the two
most challenging environments, i.e.  WAREHOUSE (fig. \ref{fig:face}; top) and
WILLOWGARAGE, whose maps are Occupancy Grids, and in which a panoramic range
sensor is placed at $16$ different poses, respectively $\bm{p}_{\ast}^{W}$ and
$\bm{p}_{\ast}^{G}$, for $N = 100$ independent attempts at global localisation
per pose. The tests are conducted with the use of a sensor whose number of rays
$N_s = 360$, maximum range $r_{\max} = 10.0$ m, and noise $\sim \mathcal{N}(0,
0.05^2)$ [m,m$^2$].  CBGL's internal \texttt{sm2} method is chosen to be
\texttt{x1} \cite{Filotheou2023a} due to the periodicity of the range signal,
\texttt{x1}'s robust pose errors compared to \texttt{sm2} state of the art
methods, and its greater ability in matching scans captured from greater
initial displacements than ICP alternatives \cite{Filotheou2023a}.  The latter
translates to the need for smaller initial hypotheses sets: for each
environment the locational density is set to $3$\texttt{e}+04 divided by the
free space area of each environment. For Monte Carlo approaches MCL and GMCL
the number of initial hypotheses is also set to $3$\texttt{e}+04.

The maximum range of the sensor is such that the geometry of environment
WAREHOUSE causes disorderly and extended lack of sampling of the sensor's
surrounding environment, which limits available information and may therefore
produce spurious measurements and increase ambiguities between the ranks of
candidate estimates. In WILLOWGARAGE, on the other hand, almost all sensor
placements result in complete sampling of its surroundings, but the sensor is
purposefully posed in such conditions as to challenge the localisation methods'
ability to perform fine distinctions between similar surroundings. Figure
\ref{fig:b:inliers_per_pose} depicts the overall distribution of position
errors, orientation errors, and execution times per tested environment and
algorithm, and fig. \ref{fig:b:boxplots} depicts the percentage of outputs
whose position error is lower than $\delta_{\bm{l}} = 0.5$ m per tested pose.
Although the cardinality of set $\mathcal{H}$ is equal in both environments,
CBGL's execution times are uneven due to \texttt{x1}'s increased execution time
when dealing with scans with missing range information. ALS is more robust
against missing information than against repeated surroundings, while PGL-PLICP
exhibits the inverse tendency. Despite the aforementioned sources of
uncertainty, CBGL manages to exhibit the overall highest number of poses whose
position error is below $\delta_{\bm{l}} = 0.5$ m and the lowest mean position
error.


\begin{figure}[tbp]
\centering
  \vspace{-0.1cm}
\raisebox{0cm}{%
  \begin{tabular}{@{}c@{}}
  \subfloat[]{%
    {\hspace{-2.2cm}
    \input{./figures/experiments/b/xyte.tex}}
    \label{fig:b:inliers_per_pose}%
    } \vspace{-0.7cm}\\
    \subfloat[\vspace{-0.1cm}]{%
    {\hspace{-1.5cm}}
    \input{./figures/experiments/b/inliers_per_pose.tex}
    \label{fig:b:boxplots}%
  }
  \end{tabular}%
  }
  \caption{\small (a) \small Distribution of position errors, orientation
           errors, and execution times per tested environment and algorithm in
           seconds for $N_s = 360$ rays. CBGL's execution time is at least
           eighteen times lower than other Monte Carlo approaches in
           WILLOWGARAGE and four times lower in WAREHOUSE. (b) Percent
           proportions of pose outputs whose position error is lower than
           $\delta_{\bm{l}} = 0.5$ m per tested environment, pose, and method.
           Overall CBGL (green) features the highest number of inlier poses.
           }
  \vspace{-0.5cm}
\end{figure}

\subsection{Simulations against environmental and \texttt{sm2} algorithmic disparity}
\label{subsec:exp_c}

The third type of test aims to inquire how the performance of CBGL scales with
respect to increasing environment area (and therefore increased number of
hypotheses), environment diversity, sensor angular range, and choice of
overlying \texttt{sm2} method. CBGL is tested once in each of $N_E =
45402$ environments, generated via the evaluation procedure of
\cite{Filotheou2023a}, which utilises five established and publicly available
benchmark datasets provided courtesy of the Department of Computer Science,
University of Freiburg \cite{datasets_link}. The environments' sizes vary
according to fig. \ref{fig:c:time_analysis} (left). Each coordinate of the
Point Cloud map of each environment is corrupted by noise
$\sim\mathcal{N}(0,0.05^2)$ [m, m$^2$].  The angular range of the range sensor
varies according to the overlying \texttt{sm2} method used: for the Normal
Distribution Transform (NDT) \cite{ndt}, Fast Generalised ICP (FastGICP)
\cite{fgi}, and Fast Voxelised Generalised ICP (FastVGICP) \cite{fvg}: $\lambda
= 3\pi/2$ rad; for \texttt{x1}: $\lambda = 2\pi$ rad. Measurement noise is
$\sim \mathcal{N}(0,0.03^2)$ [m,m$^2$].

Figure \ref{fig:c:errors_and_time} illustrates that, with the exception of NDT,
all versions of CBGL exhibit mean positional errors less than $1.0$ m; CBGL's
combination with \texttt{x1} exhibits a mean error of approximately $0.5$ m.
The evidence illustrate that CBGL is robust to sensor angular range, as the
distributions of errors between bottom-$k$ ($\mathcal{H}_1$) sets are virtually
indistinguishable for $k=10$. Figure \ref{fig:c:time_analysis} (left) shows the
execution time of CBGL combined with FastVGICP or \texttt{x1} as a function of
environmental area, and the timing breakdown of the combination of CBGL with
\texttt{x1} (right) with respect to (a) CBGL's total time minus \texttt{sm2}
time and (b) computing map-scans, as proportions of total execution time.

\begin{figure}[H]
  \input{./figures/experiments/c/errors.tex}
  \vspace{0.01cm}
  \caption{\small Distributions of positional and orientational errors and of
           execution time of CBGL for varying choices of scan--to--map-scan
           matching methods. The errors of CBGL's internal $\mathcal{H}_1$ set
           are virtually unaffected by the decrease in angular range $\lambda$
           ($\lambda_{\text{NDT}} = \lambda_{\text{FastGICP}} =
           \lambda_{\text{FastVGICP}} = 3\pi/2 \neq \lambda_{\texttt{x1}} = 2\pi$)
           }
  \label{fig:c:errors_and_time}
  \vspace{0.2cm}
\end{figure}

\begin{figure}[H]
  \input{./figures/experiments/c/time_analysis.tex}
  \vspace{0.6cm}
  \caption{\small Left: CBGL's execution time with respect to environment area
           for two choices of overlying scan--to--map-scan matching methods. In
           rough terms $\mu_t^{\text{CBGL}} \text{ [sec]} \simeq 10^{-2}\cdot area
           \text{ [} \text{m}^2 \text{]}$ for areas larger than 200 m$^2$.
           Right: Proportion of CBGL $\circ$ \texttt{x1}'s total execution time
           spent on (a) all operations up to and except for matching, and (b)
           computing map-scans, with respect to area}
  \label{fig:c:time_analysis}
\end{figure}

\section{Characterisation \& Limitations}
  \label{section:characterisation}
  Range scans with panoramic angular ranges induce fewer pose ambiguities in pose
estimate rankings than those with non-panoramic field of view. In the latter
case this means that, given the evidence of subsections \ref{subsec:exp_a} and
\ref{subsec:exp_c} where $\lambda = 3\pi/2$ rad, the choice of $k=10$ largely
inhibits the propagation of ambiguities to the output (fig.
\ref{fig:c:errors_and_time}).  However, non-panoramic sensors coupled with
repeated environment structures may give rise to the conditions of fig.
\ref{fig:h_and_h_not_fig} (bottom). Other sources of potential, large pose
errors for CBGL are portrayed in fig. \ref{fig:a:map_and_trajectory}:
Regions coloured with cyan indicate closed glass doors, wherein high range
errors in $\mathcal{S}_R$---which result from premature and arbitrary
reflections of the LIDAR's light on glass---form major discrepancies with
regard to map-derived virtual ranges. These subsequently propagate to CAER
values and ultimately corrupt the estimation of pose error hierarchies derived
from these values.  Discrepancies of this kind arise also in regions coloured
purple, which indicate vicinities around doors. In these areas the non-linear
contour of ranges, combined with the fact that from different position
estimates different parts of the environment become visible (and therefore
small changes in position may result in large discrepancies between real and
virtual scans)---these factors may cause to require higher values of locational
density $d_{\bm{l}}$ or values of $k$ in order to suppress highly erroneous
pose estimates propagated to the output.

\begin{figure}[H]\vspace{-0.2cm}
  \input{./figures/h_and_h_not_fig.tex}
  \vspace{0.3cm}
  \caption{\small Top figures: In principle there is at least one admissible
           pose estimate ($\delta \leq \delta_0$) in $\mathcal{H}$ for any
           choice of $k$. However, $k$ sets threshold $\psi_0$ on the CAER of
           estimates in $\mathcal{H}$, and therefore it sets $\mathcal{V}$,
           where $\mathcal{V} \cup \mathcal{X} \cup \mathcal{W} = \mathcal{H}$.
           In case of (a) repetitions of the immediate environment of the
           sensor in a given map, and (b) non-panoramic angular range of a
           sensor, it is possible that a choice of $k$ may starve $\mathcal{V}$
           of admissible poses, as witnessed in the two figures at the bottom
           (this is also true for pose $\bm{p}_{i}^G$ in environment
           WILLOWGARAGE of subsection \ref{subsec:exp_b}). See \cite{cbglarxiv}
           for details on $\mathcal{V}$, $\mathcal{X}$, and $\mathcal{W}$
        }
  \label{fig:h_and_h_not_fig}
\end{figure}

\section{Conclusions and Future Steps}
  \label{section:finale}
  This article has presented a single-shot Monte Carlo approach to the solution
of the passive version of the global localisation problem with the use of a 2D
LIDAR sensor, titled CBGL. CBGL allows for the fast estimation of the sensor's
pose within a metric map by first dispersing hypotheses in it and then
leveraging (a) the proportionality of values of the Cumulative Absolute Error
per Ray (CAER) metric to the position and orientation errors of the hypotheses,
for estimates in a neighbourhood of the sensor's pose, and (b) the lack of
disproportionality outside of it. CBGL was evaluated in various real and
simulated conditions and environments; it was found to be superior to Monte
Carlo and feature-based approaches in terms of number of inlier pose estimates
and execution time. Future steps will aim at (a) the extension of the CAER
metric for the use with 3D LIDAR sensors in service to a solution of the
problem of their localisation in 6DoF, and (b) making CBGL more robust by
considering the statistics of clusters in case of disparate estimates in its
$\mathcal{H}_2$ set. The C++ ROS code of the proposed method is available at
\url{https://github.com/li9i/cbgl}.

\bibliographystyle{ieeetran}
\bibliography{./sections/bibliography_ieee}


\balance

\end{document}